\pdfoutput=1

\documentclass[11pt]{article}

\usepackage{acl}

\usepackage{times}
\usepackage{latexsym}
\usepackage{amsmath}
\usepackage{siunitx}
\usepackage{algorithm} 
\usepackage{algpseudocode} 
\usepackage{mathtools}
\usepackage{graphics}
\usepackage{makecell}

\usepackage[T1]{fontenc}

\usepackage[utf8]{inputenc}

\usepackage{microtype}

\title{GRS: Combining Generation and Revision in Unsupervised Sentence Simplification}

\author{Mohammad Dehghan$^1$, Dhruv Kumar$^2$, Lukasz Golab$^1$\\
  $^1$University of Waterloo \\
  $^2$Grammarly\\
  \texttt{\{m25dehgh,lgolab\}@uwaterloo.ca} \\
  \texttt{dhruv.kumar@grammarly.com}}
  
\begin{document}
\maketitle
\begin{abstract}
We propose GRS: an unsupervised approach to sentence simplification that combines text generation and text revision.  We start with an iterative framework in which an input sentence is revised using explicit edit operations,  
and add paraphrasing 
as a new edit operation.  This allows us to combine the advantages of generative and revision-based approaches: paraphrasing captures complex edit operations,  
and the use of explicit edit operations in an iterative manner provides controllability and interpretability.  We demonstrate these advantages of GRS compared to existing methods on the Newsela and ASSET datasets.
\end{abstract}



\section{Introduction}
Text simplification is the task of reducing the complexity and improving the readability of text while preserving its meaning. This is beneficial for persons with reading disabilities~\citep{evans-etal-2014-evaluation}, non-native speakers, people with low literacy, and children.  Furthermore, other NLP tasks can use simplification as a  pre-processing step, such as 
summarization ~\citep{klebanov2004text}, parsing ~\citep{chandrasekar-etal-1996-motivations}, and machine translation~\citep{stajner-popovic-2016-text}.

Sentence simplification models can be categorized into \textit{generative} and \textit{revision-based}. Generative approaches produce a simple sentence from a complex sentence in one step, in an auto-regressive way
\citep{zhang-lapata-2017-sentence, guo-etal-2018-dynamic, kriz2019complexity, surya-etal-2019-unsupervised,  martin-etal-2020-controllable}.
Revision-based methods iteratively edit a given sentence using a sequence of edit operations such as word deletion~\citep{alva-manchego-etal-2017-learning, dong-etal-2019-editnts, kumar-etal-2020-iterative, agrawal-etal-2021-non}. While generative models learn complex edit operations implicitly from data, the explicit edit operations performed by revision-based approaches can provide more control and interpretability.

Simplification methods can also be categorized as supervised or unsupervised. Supervised methods tend to have better performance, but require aligned complex-simple sentence pairs for training~\citep{zhang-lapata-2017-sentence, guo-etal-2018-dynamic, kriz2019complexity, martin-etal-2020-controllable, martin2020muss, maddela-etal-2021-controllable}. Unsupervised methods do not need such training data but do not perform as well~\citep{surya-etal-2019-unsupervised, kumar-etal-2020-iterative, Zhao_Chen_Chen_Yu_2020}.

We propose GRS: a new approach to bridge the gap between generative and revision-based methods for unsupervised sentence simplification. 
The insight is to introduce \emph{paraphrasing} as an edit operation within an iterative revision-based framework. For paraphrasing, 
we use a fine-tuned BART model~\citep{lewis-etal-2020-bart} with lexically-constrained decoding \cite{hokamp-liu-2017-lexically, post-vilar-2018-fast, hu-etal-2019-improved}. This decoding technique allows us to select words from the initial sentence that must be changed in the paraphrased sentence  (otherwise, paraphrasing an entire sentence reduces to a pure generative model).
To avoid the computational overhead of repeatedly performing constraint-based decoding using various combinations of words to paraphrase, GRS includes a complex component detector to identify the most appropriate words to paraphrase.  
The code is available at \url{https://github.com/imohammad12/GRS}.

GRS is unsupervised in the sense that it 
does not require aligned complex-simple sentence pairs, but it uses supervised models. The paraphrasing model requires paraphrasing corpora, and the complex component detector requires two unlabeled corpora, one containing more complex sentences than the other.
However, collecting paraphrasing data and unaligned simplification data is simpler than collecting aligned complex-simple pairs. 

\section{Related Work}

Early work on simplification relied on rules, e.g., to split or shorten long sentences~\citep{CHANDRASEKAR1997183, carroll1998practical, vickrey-koller-2008-sentence}. Later work treated simplification as a monolingual phrase-based machine translation (MT) task \cite{coster-kauchak-2011-learning, wubben-etal-2012-sentence}, 
with syntactic information added, such as constituency trees \cite{zhu-etal-2010-monolingual}. 
Recent work, reviewed below, leverages neural models in a \textit{generative} and \textit{revision-based} manner. 


\textbf{Supervised Generative Methods} employ Seq2Seq 
models to learn simplification operations from aligned complex-simple sentence pairs 
\cite{nisioi-etal-2017-exploring}. 
Building on a Seq2Seq model, 
\citet{zhang-lapata-2017-sentence} used reinforcement learning to optimize a reward based on simplicity, fluency and relevance. Recent methods build on transformer~\citep{NIPS2017_3f5ee243} models, by integrating external databases containing simplification rules~\citep{zhao-etal-2018-integrating}, using an additional loss function to generate diverse outputs~\citep{kriz2019complexity}, combining syntactic rules ~\citep{maddela-etal-2021-controllable}, and conditioning on length and syntactic and lexical complexity features~\cite{martin-etal-2020-controllable}.

\textbf{Unsupervised Generative Methods} rely on non-aligned complex and simple corpora.
\citet{Zhao_Chen_Chen_Yu_2020} adopted a back-translation framework, whereas \citet{surya-etal-2019-unsupervised} used an unsupervised style transfer paradigm. \citet{martin2020muss} used a pre-trained BART model fine-tuned on paraphrased sentence pairs.

\textbf{Controllable Generative Methods} produce outputs at specified grade levels~\citep{scarton-specia-2018-learning, nishihara-etal-2019-controllable}, or apply syntactic or lexical constraints on the generated sentences~\citep{martin-etal-2020-controllable, martin2020muss}. However, these models do not provide any insights into the simplification process.

\textbf{Supervised Revision-Based Methods} use complex-simple sentence pairs to learn where to apply edit operations. \citet{alva-manchego-etal-2017-learning} use keep, replace, and delete operations. Some recent work used iterative non-autoregressive models to edit sentences by either predicting token-level edit-operations~\citep{omelianchuk-etal-2021-text} or using a fixed pipeline of edit operations~\citep{agrawal-etal-2021-non}. \citet{dong-etal-2019-editnts} proposed a hybrid method with explicit edit operations in an end-to-end generative model. 



\textbf{Unsupervised Revision-Based Methods} such as \citet{narayan2016unsupervised} apply a pipeline of edit operations in a fixed order. 
\citet{kumar-etal-2020-iterative} presented an unsupervised revision-based approach by modelling text simplification as an unsupervised search problem.  
While GRS also uses a revision-based framework and an unsupervised search strategy, we integrate a generative paraphrasing model into the framework to leverage the strengths of both text generation and text revision approaches.



\section{GRS Model}



\subsection{Overview}
\label{sec:3.1_over}
Our solution, GRS, iteratively revises a given complex sentence by applying edit operations on sentence fragments. 
In each iteration, multiple candidate simplifications are produced and evaluated using a scoring function (Section~\ref{sec:3.5_score_function}), and the best candidate is selected (Section~\ref{sec:3.6_simp_search}). The selected sentence acts as the input to the next iteration. This process continues until none of the candidate sentences are simpler than the input sentence. 

GRS uses two edit operations: 
\textit{paraphrasing} (Section~\ref{sec:3.2_par_op}; guided by the complex component detector described in Section~\ref{sec:3.3_ccd}) and \textit{deletion} (Section~\ref{sec:3.4_del_op}). 
%
%
The scoring function (Section~\ref{sec:3.5_score_function}) guides our search for best simplifcation, using soft and hard constraints on simplicity, linguistic acceptability, and meaning preservation. 

\begin{figure}[t]
\centering
\includegraphics[width=.4\textwidth]{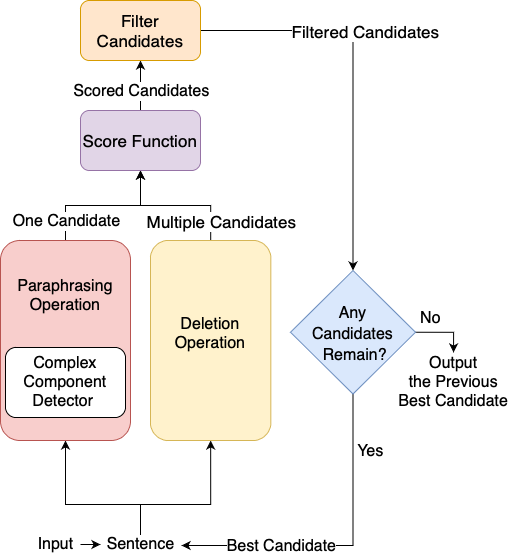} 
\caption{An overview of GRS. Given a complex input sentence, simplifications are iteratively produced via paraphrasing and deletion, with paraphrasing guided by the complex component detector. Sentences passing a filter (Equation \ref{eq:threshold}) are candidates for input to the next iteration.}
\label{fig:system_cycle}
\end{figure}

In Section~\ref{sec:3.6_simp_search}, we explain how paraphrasing and deletion work in an iterative search framework, how candidate sentences are selected, and when the algorithm terminates. Figure~\ref{fig:system_cycle} gives an overview of GRS, which is explained further in Section~\ref{sec:3.6_simp_search}.

\subsection{Paraphrasing Operation}
\label{sec:3.2_par_op}
\begin{figure}[t]
\centering
\includegraphics[width=.45\textwidth]{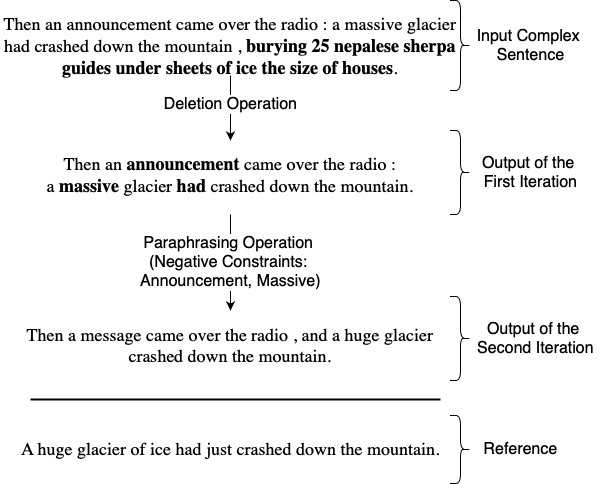} 
\caption{Two iterations of edit operations: deletion, then paraphrasing  
to simplify the complex fragments (``announcement'' and ``massive'') identified by the complex component detector 
and given as negative constraints to the paraphrasing model. This example demonstrates the interpretability of GRS through building a simplification path leading to the final sentence.}
\label{fig:example}
\end{figure}

We use a pre-trained BART model \citep{lewis-etal-2020-bart}, fine-tuned on a small subset of ParaBank 2 paraphrasing dataset \citep{hu-etal-2019-large}; however, any paraphrasing auto-regressive model can be used instead. 
During inference, we use lexical-constrained decoding \cite{hokamp-liu-2017-lexically, post-vilar-2018-fast, hu-etal-2019-improved} to place negative constraints on complex words and phrases in the input sentence. Negative constraints are words that the paraphrasing model is forced not to generate during decoding. Figure~\ref{fig:example} shows an example in which an input sentence was paraphrased to exclude two complex words (negative constraints): ``massive'' and ``announcement''.  We explain how to choose negative constraints below, with the help of the complex component detector.

\subsection{Complex Component Detector}
\label{sec:3.3_ccd}

Constrained decoding is computationally more expensive than greedy and beam search decoding. 
In GRS, before paraphrasing a sentence, the complex component detector 
predicts the best negative constraints; then the sentence and the predicted negative constraints are given to the paraphrasing model to generate a new candidate sentence.  As a result, the paraphrasing operation is called only once per iteration of GRS, using the predicted negative constraints, avoiding the expensive process of repeatedly paraphrasing the input using different combinations of negative constraints. 




We implemented the complex component detector as a complex-simple classifier that gives a simplicity probability to a given sentence. We only require two corpora with different complexity levels to train this classifier. Since aligned complex-simple sentence pairs are not required, this classifier can be trained on any pair of corpora with different complexity levels. 
\citet{reid-zhong-2021-lewis} showed that it is possible to extract style-specific sections of a sentence using the attention layers of a style classifier. 
Similarly, we use the attention layers of our complex-simple classifier to extract the complex components from a given input sentence. 

\begin{figure}[t]
\centering
\includegraphics[width=0.24\textwidth]{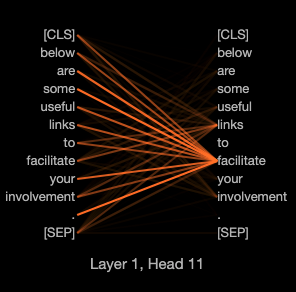} 
\caption{One of the attention matrices of the DeBERTa complex-simple classifier (head 11 of the second layer). Attention weights are reflected by color intensity. The input sentence in this example is ``below are some useful links to facilitate your involvement." 
We used BertViz~\cite{vig-2019-multiscale} to visualize attention weights. 
}
\label{fig:att}
\end{figure}

We fine-tune the pre-trained DeBERTa model~\citep{he2021deberta} as our complex-simple classifier. Figure \ref{fig:att} illustrates one of the attention heads of the second layer of DeBERTa. This visualization shows that the word ``faciliate'' was attended to more than the other words in the given sentence. We use this intuition and devise a formula (Equations~\ref{eq:attention-1} and \ref{eq:attention-2} below) to detect complex words by analyzing attention weights. 


BERT~\citep{devlin-etal-2019-bert} and its extensions (e.g. DeBERTa) add a [CLS] token to the beginning and a [SEP] token to the end of each sentence (as shown in Figure \ref{fig:att}). In these models, the hidden states of the [CLS] token in the last layer are used for classification tasks. In our complex-simple classifier, we found that the attention paid by the [CLS] token in the second layer to other words in the sentence can help us detect complex components. Equations \ref{eq:attention-1} and \ref{eq:attention-2} demonstrate how we extract complex components from attention head matrices of the second layer of the classifier. Here, $\mathbf{A}_{h, i}^{[CLS]}$ refers to the amount of attention the [CLS] token in the $h$th attention head of the second layer pays to the $i$th token of the input sentence. $N$ and $H$ refer to the length of the input sentence and the number of attention heads, respectively. $c_{i}$ defines whether the $i$th token is complex or not. If $c_{i}=1$, then this token will be set as a negative constraint. $\bar{T}$ is a threshold used for finding complex tokens. In the example demonstrated in Figure \ref{fig:att}, only $c_{8}$, which refers to the word ``facilitate", is a complex token.

\begin{equation}
\label{eq:attention-1}
\begin{aligned}
\bar{T} =  \frac{ \sum_{h=0}^{H-1}\sum_{i=0}^{N-1}\mathbf{A}_{h, i}^{[CLS]}}{N}
\end{aligned}
\end{equation}

\begin{equation}
\label{eq:attention-2}
\begin{aligned}
   c_{i}  = 
\begin{dcases}
    1, & \text{if  }  \sum_{h=0}^{H-1}\mathbf{A}_{h, i}^{[CLS]
    }   \geq  \bar{T}\\ 
    0, & \text{otherwise}
\end{dcases}
\end{aligned}
\end{equation}


\subsection{Deletion Operation}
\label{sec:3.4_del_op}
Deletion aims to remove peripheral information to make sentences simpler, and is composed of two sub-operations: removal and extraction. Inspired by ~\citet{kumar-etal-2020-iterative}, we use the \emph{constituency tree} of the input sentence to obtain all constituents from different depths of the parse tree. These constituents can be deleted (removal) or selected as a simplified candidate sentence (extraction). The removal sub-operation creates new candidate sentences by removing each of these phrases from the input sentence. The extraction sub-operation selects phrases as candidate sentences, which helps the model extract the main clause and drop peripheral information.  The example in Figure \ref{fig:example} drops the phrase \textit{``burying 25 nepalese sherpa guides under sheets of ice the size of houses''} from the complex sentence since it is not the main clause. 

\subsection{Scoring Function}

Candidates generated by our two edit operations may not be correct in terms of linguistic acceptability. Furthermore, important information from the original sentence may have been removed. We use a score function to filter out non-grammatical candidates or sentences that are not conceptually similar to the original sentence. The score function is composed of three important components. 

\textbf{Meaning Preservation ($H_{mp}$):}
\label{sec:3.5_score_function}
First, we use the method proposed in \citet{reimers-2019-sentence-bert} to obtain the semantic representations of the sentences. We then use the cosine similarity measure between the representations of the original and the generated candidate sentence. Our meaning preservation measure acts as a hard filter. A hard filter assigns a zero score to candidate sentences that do not pass a certain threshold. 

\textbf{Linguistic Acceptability ($H_{la}$):}
By removing some components of a complex input sentence, the output sentence may become nonsensical. To check the linguistic acceptability of the generated sentences, we train a classifier on the CoLA (the corpus of linguistic acceptability) ~\citep{warstadt-etal-2019-neural} dataset. This classifier measures the probability that a given sentence is grammatical. This module, like the meaning preservation module, is used as a hard filter in the score function.

\textbf{Simplicity ($S_{simp}$):}
This is a soft constraint,  
for which we use the complex-simple classifier mentioned in Section \ref{sec:3.3_ccd}, which computes the simplicity probability of a sentence.

These three measures together evaluate the quality of each candidate sentence, as shown in Equation \ref{eq:score_function}. In this equation, $S$, $S_{simp}$, $H_{la}$, $H_{mp}$, $c$, and $o$ refer to the score function, simplicity module, linguistic acceptability hard filter, meaning preservation hard filter, candidate sentence, and the original sentence, respectively. 

\begin{equation}
\label{eq:score_function}
\begin{aligned}
S(c) =  S_{simp}(c) * H_{mp}(c, o)  * H_{la}(c) 
\end{aligned}
\end{equation}



\subsection{Simplification Search}
\label{sec:3.6_simp_search}
Our unsupervised search method is inspired by \citet{kumar-etal-2020-iterative}, but with different simplification operations and a different score function. Given a complex input sentence, paraphrasing and deletion operations generate candidate sentences separately. In each iteration, the paraphrasing operation creates only one candidate sentence, as described in Section~\ref{sec:3.2_par_op}, whereas the deletion operation generates multiple candidate sentences (Section~\ref{sec:3.4_del_op}). Candidates sentences are then evaluated according to the scoring function. Given a score for each candidate sentence, we filter out those candidates that do not improve the score of the input sentence by some threshold. The threshold depends on the edit operation that the candidate sentence has been created from. In equation \ref{eq:threshold}, $t_{op}$ is the threshold associated with operator $op$. $S$, $c$, and $c^{\prime}$ refer to the score function, the candidate sentence, and the input sentence in the current iteration, respectively. 


\begin{equation}
\label{eq:threshold}
\begin{aligned}
S(c) > S(c^{\prime}) * t_{op}
\end{aligned}
\end{equation}

Finally, at the end of each iteration, out of the remaining sentences (that are not filtered out), we select the one with the highest score. 

\section{Experiments}

This section discusses the experimental setup (Sections~\ref{sec:data} through \ref{subsec:models_tested}), a comparison of GRS with existing approaches (Section~\ref{sec:4.5_eval_restuls}), a controllability study (Section~\ref{sec:cont}), an evaluation of the complex component detector (Section~\ref{sec:4.7_ccd_eval}), an analysis of the simplification search (Section~\ref{sec:4.8_search_eval}), and a human evaluation  (Section~\ref{sec:4.9_human_eval}). 

\subsection{Data}
\label{sec:data}
We use the Newsela ~\citep{xu-etal-2015-problems} and ASSET datasets~\citep{alva-manchego-etal-2020-asset} to evaluate GRS against existing simplification methods.
Newsela contains 1840 news articles for children at five reading levels. We use the split from \citet{zhang-lapata-2017-sentence}, containing 1129 validation and 1077 test sentence pairs.
ASSET includes 2000 validation and 359 test sentences pairs. Each sentence has ten human-written references. 

\begin{table*}[t]
\centering
\resizebox{12cm}{!}{%
\begin{tabular}{lllllll}
\hline
\textbf{Model} & \textbf{SARI $\uparrow$} & \textbf{Add $\uparrow$} & \textbf{Delete $\uparrow$} & \textbf{Keep $\uparrow$} & \textbf{FKGL $\downarrow$} & \textbf{Len}\\
\hline
Identity Baseline & 12.24 &  0.00 & 0.00 & 36.72 & 8.82 & 23.04 \\
\hline
\textbf{Unsupervised Models} & & \\
\hline
\citet{Zhao_Chen_Chen_Yu_2020} & 37.20 & 1.51 & 73.53 & 36.54 & 3.80 & 11.78 \\ 
\citet{kumar-etal-2020-iterative} & 38.36 & 1.01 & 77.58 & 36.51 & \textbf{2.81} & 9.61 \\ 
\citet{martin2020muss} & 38.29 & \textbf{4.44} & 76.02 & 34.42 & 4.65 & 12.49 \\ 
GRS: DL (RM + EX) & 37.52 & 0.66 & 69.45 & \textbf{42.44} & 3.93 & 12.64 \\
GRS: PA & 36.42 & 3.44 & 69.55 & 36.28 & 5.79 & 19.08 \\
GRS: PA + DL (RM + EX) & \textbf{40.01} & 3.06 & \textbf{80.43} & 36.53 & 3.20 & 11.72 \\

\hline
\textbf{Supervised Models} & & \\
\hline
\citet{narayan-gardent-2014-hybrid} & 34.73 & 0.77 & 73.22 & 30.21 & 4.52 & 12.40 \\ 
\citet{zhang-lapata-2017-sentence} & 38.03 & 2.43 & 69.47 & \textbf{42.20} & 4.78 &  14.36 \\ 
\citet{dong-etal-2019-editnts} & 39.28 & 2.13 & 77.17 & 38.53 & 3.80 & 10.92 \\ 
\citet{Zhao_Chen_Chen_Yu_2020} & 39.14 & 2.80 & 74.28 & 40.34 & 4.11 & 11.63 \\ 
\citet{martin2020muss} & \textbf{41.20} & \textbf{6.02} & \textbf{81.70} & 35.88 & \textbf{2.35} & 9.22 \\ 

\hline
\end{tabular} %
}
\caption{\label{tab:newsela}
Comparison of supervised and unsupervised simplification models on the Newsela test set. PA and DL refer to paraphrasing and deletion, respectively.  RM and EX refer to the removal and extraction sub-operations of the deletion operation.  $\uparrow$ denotes the higher the value, the better. $\downarrow$ denotes the lower the value, the better.}
\end{table*}

\subsection{Training Details}

\textbf{Paraphrasing Model:} 
We fine-tune a pre-trained BART model~\citep{lewis-etal-2020-bart} implemented by \citet{wolf-etal-2020-transformers}. To do this, we use a subset of the ParaBank 2 paraphrasing dataset \citep{hu-etal-2019-large} containing 47,000 pairs.  
We observe that a conservative paraphrasing model helps us to control the generated output. This is because such a model is better at specifically changing only words provided as negative constraints.
Thus, for fine-tuning the BART model, we select paraphrasing sentence pairs that are semantically similar to each other. 
For calculating the semantic similarity of sentence pairs, we use the model from \citet{reimers-2019-sentence-bert} to obtain sentence embeddings, and then we use cosine-similarity to find the most similar sentence pairs. 

The BART model is composed of a 12-layer encoder and a 12-layer decoder, each layer containing 16 attention heads. The model's hidden size is 1024, and the tokenizer vocabulary size is 50265. We use a batch size of 8 (per device). It took approximately 1.5 hours to fine-tune the model using three NVIDIA 2080 Ti GPUs.


\textbf{Complex-Simple Classifier:} 
We use a pre-trained DeBERTa~\citep{he2021deberta} model implemented by \citet{wolf-etal-2020-transformers}. This model is composed of a 12-layer self-attentional encoder, each layer containing 12 attention heads. The model's hidden size is 768, and the tokenizer vocabulary size is 30522. To fine-tune the DeBERTa model for the binary classification task, we use the Newsela-Auto dataset~\citep{jiang-etal-2020-neural}. 
To train the classifier, we use the AdamW~\citep{loshchilov2018decoupled} optimizer with a learning rate of \num{5e-5} and a batch size of 16. Note that we do not use the alignment between complex-simple sentence pairs in the Newsela-Auto dataset. Thus, our complex-simple classifier can be trained on any text corpora with different complexity levels.  It took approximately one hour to fine-tune the classifier using a single NVIDIA 2080 Ti GPU. The accuracy of this classifier is 78.46.


\textbf{Meaning Preservation Module of the Scoring Function:}
To obtain contextual embeddings of sentences, we use the SentenceTransformers \cite{reimers-2019-sentence-bert} framework, specifically, the paraphrase-mpnet-base-v2 pre-trained model. 

\textbf{Linguistic Acceptability Module of the Scoring Function:}
To score the linguistic acceptability of a sentence, we fine-tune a pre-trained DeBERTa model~\citep{he2021deberta} for a binary classification task on the CoLA (the corpus of linguistic acceptability) ~\citep{warstadt-etal-2019-neural} dataset. It contains 10,657 sentences, each labelled either as grammatical or ungrammatical. The configuration and training hyperparameters of this classifier are the same as the complex-simple classifier explained above.  It took approximately 30 minutes to fine-tune the model using a single NVIDIA 2080 Ti GPU. On the validation set, the accuracy of the model is 79.33.

\begin{table*}[t]
\centering
\resizebox{12cm}{!}{
\begin{tabular}{lllllll}
\hline
\textbf{Model} & \textbf{SARI $\uparrow$} & \textbf{Add $\uparrow$} & \textbf{Delete $\uparrow$} & \textbf{Keep $\uparrow$} & \textbf{FKGL $\downarrow$} & \textbf{Len} \\
\hline
Identity Baseline & 20.73 &  0.00 & 0.00 & 62.20 & 10.02 & 19.72\\
Gold Reference & 44.89 & 10.17 & 58.76 & 65.73 & 6.49 & 16.54 \\
\hline
\textbf{Unsupervised Models} & & \\
\hline

\citet{surya-etal-2019-unsupervised} & 35.19 & 0.83 & 45.98 & 58.75 & 7.60 & 16.81  \\ 
\citet{Zhao_Chen_Chen_Yu_2020} & 33.95 & 1.99 & 42.09 & 57.77 & 7.51 & 18.80 \\ 
\citet{kumar-etal-2020-iterative} & 36.67 & 1.29 & 51.33 & 57.40 & 7.33 & 16.56\\ 
\citet{martin2020muss} & \textbf{42.42} & \textbf{7.15} & 61.32 & \textbf{58.77} & 7.49 & 16.36 \\ 
GRS: DL (RM + EX) & 37.90 & 0.89 & 62.32 & 50.50 & 4.17 & 11.18 \\
GRS: PA & 40.41 & 7.00 & 62.37 & 51.88 & 6.70 & 17.94 \\
GRS: PA + DL (RM + EX) & 37.40 & 3.89 & \textbf{67.46} & 40.85 & \textbf{3.45} & 10.69 \\

\hline
\textbf{Supervised Models} & & \\
\hline
\citet{narayan-gardent-2014-hybrid} & 34.65 & 1.3 & 59.24 & 43.41 & \textbf{5.18} & 10.95 \\ 
\citet{xu-etal-2016-optimizing} & 37.11 & 5.07 & 45.21 & 61.06 & 7.95 &  20.50 \\ 
\citet{zhang-lapata-2017-sentence} & 36.59 & 2.38 & 50.10 & 57.30 & 7.66 & 14.37\\ 
\citet{zhao-etal-2018-integrating} & 38.67 & 4.36 & 51.37 & 60.29 & 7.73 & 18.36 \\ 
\citet{dong-etal-2019-editnts} & 34.95 & 2.40 & 42.69 & 59.73 & 8.38 & 16.49 \\ 
\citet{martin-etal-2020-controllable} & 40.13 & 6.53 & 50.84 & \textbf{62.99} & 7.29 & 19.49 \\ 
\citet{Zhao_Chen_Chen_Yu_2020} & 35.15 & 2.22 & 45.32 & 57.91 & 7.83 & 16.14 \\ 
\citet{martin2020muss} & \textbf{44.05} & \textbf{10.93} & 61.91 & 59.30 & 6.13 & 18.49 \\ 
\citet{omelianchuk-etal-2021-text} & 43.21 & 8.04 & \textbf{64.25} & 57.35 & 6.87 & ------- \\ 
\hline
\end{tabular}%
}
\caption{\label{tab:asset}
Comparison of supervised and unsupervised simplification models on the ASSET test set. PA and DL refer to paraphrasing and deletion, respectively.  RM and EX refer to the removal and extraction sub-operations, the sub-operations of the deletion operation.  $\uparrow$ denotes the higher value, the better. $\downarrow$ denotes the lower value, the better.}

\end{table*}


\textbf{Simplification Search and Score Function:}
The threshold associated with paraphrasing ($t_{par}$) is 0.8, and the thresholds related to the removal ($t_{dl-rm}$) and extraction ($t_{dl-ex}$) sub-operations of the deletion operation are 1.1, and 1.25, respectively. The score function's meaning preservation ($H_{mp}$) and linguistic acceptability ($H_{la}$) thresholds 
are 0.7 and 0.3, respectively. We obtained these values using the validation set. These values are used for both ASSET and Newsela datasets.


\subsection{Evaluation Metrics}
\label{subsec:metrics} 
 
To evaluate GRS and other models, we use SARI~\citep{xu-etal-2016-optimizing} as our primary metric. 
SARI (System output Against References and against the Input sentence) evaluates the quality of the output text by calculating how often the output text correctly keeps, inserts, and deletes n-grams from the complex sentence, compared to the reference text, where 1 $\le$ n $\le$ 4.
We report the overall SARI score, as well as individual SARI scores corresponding to n-grams correctly added (ADD), deleted (DELETE) and kept (KEEP); the overall SARI score is the mean of these three scores. We also report the FKGL score, which only considers the output sentence, not the source and reference sentences. It is computed based on sentence length and the number of syllables for each word in the sentence. We use the EASSE package~\citep{alva-manchego-etal-2019-easse} to calculate SARI and FKGL. We do not use the BLEU ~\citep{papineni-etal-2002-bleu} metric since \citet{sulem-etal-2018-semantic} showed that BLEU does not correlate well with simplicity. 



\subsection{Models Tested}
\label{subsec:models_tested}

We evaluate GRS with different configurations: only deletion - GRS: DL(RM+EX), only paraphrasing - GRS: PA, and both deletion and paraphrasing - GRS: PA+DL(RM+EX). 
We also consider the complex sentence itself as a trivial baseline, denoted by `Identity Baseline'. The ASSET dataset contains multiple references for a sentence, so we also calculate an upper bound for a given evaluation metric, which we denote as `Gold Reference'. To calculate the `Gold Reference' score, each reference is selected once, and the scores are calculated against others. Finally, we average across all the reference scores to obtain the final `Gold Reference' score. 

We also compare GRS with existing approaches.  From unsupervised methods, we select unsupervised generative models that use Seq2Seq models~\citet{surya-etal-2019-unsupervised, Zhao_Chen_Chen_Yu_2020}.  We also compare with \citet{martin2020muss}, which leverages pretrained language models and a large paraphrase pair dataset, and \citet{kumar-etal-2020-iterative}, an iterative revision-based method with several explicit edit operations (deletion, lexical substitution and reordering). 

From supervised methods, we start with \citet{narayan-gardent-2014-hybrid} and \citet{xu-etal-2016-optimizing},  which use phrase-based MT models. We also consider Seq2Seq generative methods: \citet{zhang-lapata-2017-sentence}, which uses reinforcement learning, and \citet{Zhao_Chen_Chen_Yu_2020, martin-etal-2020-controllable, martin2020muss}, which use Seq2Seq transformer models. Next, we select \citet{omelianchuk-etal-2021-text}, a recent supervised revision-based method. Finally, we consider \citet{dong-etal-2019-editnts}, a hybrid approach using explicit edit operations in a generative framework.

\begin{table}[t]
\centering
\resizebox{\columnwidth}{!}{%
\begin{tabular}{l|l|l|l|l|l|l}
\hline
\textbf{Value} & \textbf{SARI $\uparrow$} & \textbf{Add $\uparrow$} & \textbf{Delete $\uparrow$} & \textbf{Keep $\uparrow$} & \textbf{FKGL $\downarrow$} & \textbf{Len} \\
\hline
\hline

\multicolumn{7}{c}{ Effect of Meaning Preservation Threshold ($H_{mp}$) } \\
\hline
\hline
0.25 & 38.18 & 2.15 & 84.64 & 27.76 & 0.42 & 7.68 \\
0.5 & 39.49 & 2.30 & 83.58 & 32.59 & 1.63 & 8.99 \\
0.6 & 39.78 & 2.59 & 81.94 & 34.80 & 2.46 & 10.29 \\
0.7 & 39.99 & 3.16 & 79.81 & 36.99 & 3.27 & 12.16 \\

\hline
\hline
\multicolumn{7}{c}{ Effect of the  Removal Threshold of Deletion Operation ($t_{dl-rm}$)} \\
\hline
\hline
0.9 & 37.33 & 1.93 & 82.72 & 27.34 & 2.19 & 8.58 \\
1.0 & 38.03 & 2.20 & 81.63 & 30.25 & 2.53 & 9.80 \\
1.1 & 40.01 & 3.06 & 80.43 & 36.53 & 3.20 & 11.72 \\
1.2 & 39.98 & 3.15 & 79.96 & 36.85 & 3.26 & 12.07 \\
\hline
\hline

\multicolumn{7}{c}{ Effect of the Paraphrasing Threshold ($t_{par}$)} \\
\hline
\hline
0.8 & 39.99 & 3.16 & 79.81 & 36.99 & 3.27 & 12.16 \\
0.9 & 40.01 & 3.15 & 79.55 & 37.31 & 3.32 & 12.24 \\
1.0 & 39.42 & 2.69 & 75.03 & 40.54 & 3.84 & 12.99 \\
1.1 & 38.55 & 1.97 & 70.47 & 43.23 & 4.11 & 13.50 \\

\hline
\hline

\multicolumn{7}{c}{ Effect of the Linguistic Acceptability Threshold ($H_{la}$)} \\
\hline
\hline
0.6 & 39.42 & 3.06 & 78.19 & 37.00 & 3.65 & 12.54 \\
0.7 & 39.52 & 3.00 & 77.98 & 37.57 & 3.68 & 12.67 \\
0.8 & 39.69 & 3.08 & 77.60 & 38.40 & 3.79 & 12.85 \\
0.9 & 38.41 & 2.93 & 76.87 & 38.42 & 4.04 & 13.04 \\


\hline
\end{tabular} %
}
\caption{\label{tab:cont}
Impact of paraphrasing, deletion, meaning preservation, and linguistic acceptability thresholds on the Newsela dataset.
}
\end{table}

\subsection{Evaluation Results}
\label{sec:4.5_eval_restuls}

Tables \ref{tab:newsela} and \ref{tab:asset}  illustrate the results on Newsela and ASSET, respectively.  We report the overall SARI score, the individual scores of three operations used in SARI score, the FKGL score, and the average length of the output sentences.  To evaluate previous methods, we obtained their output sentences on ASSET and Newsela from the respective project Github pages or by contacting the respective authors, followed by calculating the SARI and FKGL scores using the EASSE package (described in Section~\ref{subsec:metrics}).  One exception is \citet{omelianchuk-etal-2021-text}:  since they also used the EASSE package, we copied their reported ASSET scores in Table~\ref{tab:asset}, but they did not report the average sentence length.

For Newsela, using paraphrasing and deletion together (GRS: PA + DL(RM+EX)) gives the best performance on the SARI metric. On the Newsela dataset, our best model outperforms previous unsupervised methods and achieves +1.6 SARI improvement. It also outperforms all supervised methods except  \citet{martin2020muss}. 

For ASSET, even though \citet{martin2020muss} perform better than our best model, we improve the performance over \citet{kumar-etal-2020-iterative} by +3.6 SARI points and close the gap between revision-based and generative approaches. Compared to supervised models, our unsupervised model again outperforms others except \citet{martin2020muss} and \citet{omelianchuk-etal-2021-text}. For the ASSET dataset, we observe that our model with only paraphrasing (GRS (PA)) has the best SARI score.

Analyzing the results, we observe that simplification is done differently by human annotators in Newsela than in ASSET. In Newsela, removal of peripheral information through content deletion happens more aggressively. The average reference sentence length is 12.75 compared to 23.04 for the source sentences. 
However, in ASSET, content removal is conservative and can be handled by paraphrasing alone. The average reference sentence length is 16.54 compared to 19.72 for the source sentences. Simplifications in ASSET focus on lexical simplification, sentence splitting and word reordering.

\citet{martin2020muss} leverage a pretrained BART model~\citep{lewis-etal-2020-bart} and fine-tune it on a paraphrasing dataset containing 1.1 million sequence pairs. Unlike traditional paraphrasing datasets that are structured at the sentence level, their paraphrasing dataset contains multiple sentences in a sequence, thus allowing the model to learn a sentence splitting operation as well. Thus, they outperform the previous best unsupervised models on ASSET. 
On Newsela, both GRS and the model from \citet{kumar-etal-2020-iterative} perform better than \citet{martin2020muss} since they include an explicit removal edit operation. 
\citet{martin2020muss} instead do not explicitly perform content removal and only do content deletion by way of paraphrasing. Finally, \citet{kumar-etal-2020-iterative} does not perform well on ASSET since they do not perform paraphrasing. Our new design thus combines the advantages of both revision-based and generative approaches. 


\begin{table}[t]
\centering
\resizebox{\columnwidth}{!}{
\begin{tabular}{lllll}
\hline
\textbf{CCD-module} & \textbf{Acc $\uparrow$} & \textbf{Rec $\uparrow$} & \textbf{Prec $\uparrow$} & \textbf{F1 $\uparrow$} \\ 
\hline
LS-CCD & 84.67 & 37.36 & 70.51 & 48.84 \\ 
Att-Cls & 84.58 & 47.95 & 72.59 & 57.75 \\ 

\hline
\end{tabular}%
}
\caption{\label{tab:ccd-perf}
Performance of different Complex Component Detectors (CCD) on the Complex Word Identification (CWI) task. CWIG3G2 dataset has been used for this evaluation. LS-CCD and Att-Cls refer to the CCD module obtained from Lexical Simplification edit operation of \citet{kumar-etal-2020-iterative} and the original CCD module used in GRS design explained in Section~\ref{sec:3.3_ccd}, respectively. $\uparrow$ denotes the higher value, the better.}
\end{table}

\subsection{Controllability}
\label{sec:cont}

By manipulating the thresholds for the components of the score function and the edit operations, we can control the amount of deletion, paraphrasing, and the trade-off between simplicity and meaning preservation. We show the results in Table \ref{tab:cont} using the GRS (PA + DL) model and the Newsela test set. The column labels have the same meaning as in Tables~\ref{tab:newsela} and ~\ref{tab:asset}. 

\textbf{Meaning Preservation Threshold: } As mentioned in Section \ref{sec:3.5_score_function}, meaning preservation is a hard filter in our score function. 
As the meaning preservation threshold increases, candidate sentences less similar to the original sentence are pruned. Sentences more similar to the original sentence have higher Keep and lower Delete SARI scores. The SARI Add score increases since paraphrasing is prioritized over deletion. Finally, the length of the output sentences increases since the model becomes more conservative.


\textbf{Removal Threshold of Deletion Operation:} 
By increasing this threshold, the SARI Keep score increases and the SARI Delete score decreases, which also results in increased average length. 
The SARI Add score increases as well since a higher deletion threshold makes the model conservative on deletions and thus candidates from the paraphrasing operation are more likely to be selected.

\textbf{Paraphrasing Threshold:} 
Reducing this threshold results in more aggressive paraphrasing. Thus, we observe an increase in the SARI Delete and Add scores since paraphrasing replaces complex words and phrases with simpler ones. 


\textbf{Linguistic Acceptability Threshold: } Like meaning preservation, linguistic acceptability is a hard filter in our score function (Section \ref{sec:3.5_score_function}). 
As the linguistic acceptability threshold increases, more candidate sentences receive a zero score. This results in a more conservative model that makes fewer changes to the input sentences because the original sentences are already linguistically acceptable. 
By increasing the linguistic acceptability threshold, the SARI Deletion score drops and the SARI Keep score increases. Also, this results in longer sentences.


\begin{table*}[t]
\centering
\resizebox{9.5cm}{!}{%
\begin{tabular}{lllllll}
\Xhline{3\arrayrulewidth}
\textbf{Model} & \textbf{SARI $\uparrow$} & \textbf{Add $\uparrow$} & \textbf{Delete $\uparrow$} & \textbf{Keep $\uparrow$} & \textbf{FKGL $\downarrow$} & \textbf{Len} \\

\Xhline{3\arrayrulewidth}
\multicolumn{7}{c}{\textbf{ASSET}} \\ 
\hline
Identity Baseline & 20.73 &  0.00 & 0.00 & 62.20 & 10.02 & 19.72 \\
Gold Reference & 44.89 & 10.17 & 58.76 & 65.73 & 6.49 & 16.54 \\
\hline


GRS(PA, CCD:LS-CCD) & 37.80 & 5.59 & 57.39 & 50.44 & 7.17 & 18.75 \\ 
GRS(PA, CCD:Att-Cls) & \textbf{40.41} & \textbf{7.00} & \textbf{62.37} & \textbf{51.88} & \textbf{6.70} & 17.94 \\ 

\Xhline{3\arrayrulewidth}
\multicolumn{7}{c}{\textbf{Newsela}} \\ 
\hline
Identity Baseline & 12.24 &  0.00 & 0.00 & 36.72 & 8.82 & 23.04 \\
\hline

GRS(PA+DL(RM), & 39.30 & 2.87 & 78.18 & \textbf{36.85} & \textbf{3.39} & 13.54 \\ 
CCD:LS-CCD) & & & & & & \\ 
GRS(PA+DL(RM), & \textbf{39.61} & \textbf{3.18} & \textbf{79.13} & 36.52 & 3.45 & 13.43 \\ 
CCD:Att-Cls) & & & & & & \\ 

\hline
\end{tabular}%
}
\caption{\label{tab:grs-ccd}
Comparison of GRS versions that use different Complex Component Detectors (CCD) on ASSET and Newsela. PA and DL refer to paraphrasing and deletion, respectively. RM refers to removal, which is the sub-operation used in deletion operation. $\uparrow$ denotes the higher value, the better. $\downarrow$ denotes the lower value, the better.}

\end{table*}


\subsection{Complex Component Detector Evaluation}
\label{sec:4.7_ccd_eval}

To show the effectiveness of the proposed Complex Component Detector (CCD) mentioned in Section~\ref{sec:3.3_ccd}, we evaluate the CCD module on the Complex Word Identification (CWI) task. The task is defined as a sequence tagging problem in which each word in a sentence is tagged as complex or not complex. 
We use the test set of CWIG3G2 ~\citep{CWIG3G2}, a professionally written news dataset.
As explained in Section~\ref{sec:3.3_ccd}, the CCD module used in GRS (denoted by Att-Cls) operates by interpreting the attention matrix of the second layer of the complex-simple classifier. We also compare with the lexical simplification operation of \citet{kumar-etal-2020-iterative},  
denoted by LS-CCD. It uses the IDF scores to find complex words in a sentence. \\ 

Table \ref{tab:ccd-perf} shows the complex word identification performance of the two CCD modules on CWIG3G2. Att-Cls outperforms LS-CCD in recall, precision, and the F1 score. Tables \ref{tab:grs-ccd} demonstrates the performance of GRS with different CCD modules on ASSET \cite{alva-manchego-etal-2020-asset} and Newsela \cite{xu-etal-2015-problems} test sets. 
On both datasets, the GRS model using Att-Cls has higher deletion and addition scores compared to the GRS model using LS-CCD. The overall SARI score is considerably higher when using Att-Cls on ASSET.

\subsection{Simplification Search Analysis}
\label{sec:4.8_search_eval}
GRS is an interpretable unsupervised simplification method in which we can trace the simplification process. That is, we know which edit operation is applied on a given complex sentence in each iteration. Table~\ref{tab:simp-search} demonstrates how many simplification iterations were needed to simplify a complex sentence, on average, in the Newsela and ASSET datasets. We also show the average frequency of each operation to simplify a given sentence.

Table~\ref{tab:simp-search} illustrates that when both edit operations are allowed (GRS:PA+DL), almost four simplification iterations are applied to a sentence, and paraphrasing is generally more common than deletion.  


\begin{table}[t]
\centering
\resizebox{\columnwidth}{!}{
\begin{tabular}{l|c|c|c|c}
\Xhline{3\arrayrulewidth}
\textbf{Model} & \textbf{Iterations/Sent} & \textbf{PA-iterations} & \multicolumn{2}{c}{\textbf{DL-iterations}}  \\
& (all-Operations) & & RM & EX \\

\Xhline{3\arrayrulewidth}
\multicolumn{5}{c}{\textbf{Newsela}} \\ 
\hline
GRS: PA & 4.72 & 4.72 & -- & -- \\
GRS: DL & 0.79 & -- & 0.46 & 0.33 \\
GRS: PA + DL & 4.40 & 3.72 & 0.37 & 0.31 \\

\Xhline{3\arrayrulewidth}
\multicolumn{5}{c}{\textbf{ASSET}} \\ 
\hline
GRS: PA & 4.18 & 4.18 & -- & -- \\
GRS: DL & 1.05 & -- & 0.54 & 0.51 \\
GRS: PA + DL & 3.79 & 2.94 & 0.39 & 0.45 \\
\hline

\hline
\end{tabular}%
}
\caption{\label{tab:simp-search}
Analysis of edit operation used during simplification search, showing the average number of simplification iterations of GRS and the average share of each edit operation. PA and DL refer to paraphrasing and deletion, respectively.  RM and EX refer to the removal and extraction sub-operations of the deletion operation.  }

\end{table}

\subsection{Human Evaluation}
\label{sec:4.9_human_eval}

We selected 30 sentences from the ASSET test set for human evaluation. Following \cite{kriz2019complexity}, we measure Fluency (whether the sentence is grammatical and well-formed), Simplicity (whether it is simpler than the complex sentence), and Adequacy (whether it keeps the meaning of the complex sentence). We asked four volunteers to assess the sentences based on these metrics and evaluate the performance of various models, including GRS. Results 
are shown in Table \ref{tab:human-eval}. All models are unsupervised except~\citet{zhang-lapata-2017-sentence}.

The fourth column in Table \ref{tab:human-eval} shows the average score of all three metrics used in the human evaluation. According to the average scores, MUSS~\citep{martin2020muss} has the best performance, followed by GRS. The human evaluation demonstrates that the automatic evaluation (SARI scores shown in Table~\ref{tab:asset}) is aligned with human evaluation scores. 
GRS has the best performance in meaning preservation (Adequacy). This may be because we have a relatively conservative paraphrasing model. Also, GRS evaluated in this study is only leveraging paraphrasing, and this version is the most conservative. 

\begin{table}[t]
\centering
\resizebox{\columnwidth}{!}{
\begin{tabular}{l|c|c|c|c}
\hline
\textbf{CCD-module} & \textbf{Adequacy $\uparrow$} & \textbf{Simplicity $\uparrow$} & \textbf{Fluency $\uparrow$} & \textbf{Average $\uparrow$} \\ 
\hline
Reference   &  4.29 &  4.08 &  4.76 &  4.37 \\ 
\hline
GRS(PA)        &  \textbf{3.98} &  4.04 &  4.47 &  4.17 \\ 
\cite{surya-etal-2019-unsupervised}       &  3.57 &  3.48 &  4.32 &  3.79 \\ 
\cite{Zhao_Chen_Chen_Yu_2020}     &  3.89 &  3.27 &  4.54 &  3.89 \\ 
\cite{martin2020muss}   &  3.95 &  \textbf{4.17} &  \textbf{4.78} &  \textbf{4.30} \\ 
\cite{kumar-etal-2020-iterative} &  3.15 &  3.56 &  4.15 &  3.62 \\ 
\cite{zhang-lapata-2017-sentence}  &  3.67 &  3.64 &  4.69 &  4.00 \\ 

\hline
\end{tabular}%
}
\caption{\label{tab:human-eval}
Human evaluation on the ASSET dataset. Adequacy, simplicity, and fluency are human evaluation metrics, and in the fourth column, the average of these metrics is shown. Each row represents a simplification model. Human evaluation scores are based on a 1–5 Likert scale. $\uparrow$ denotes the higher value, the better.}
\end{table}

\section{Conclusion}
We proposed GRS, a controllable and interpretable method for unsupervised text simplification that bridges the gap between previous unsupervised generative and revision-based approaches. We combined the two approaches by incorporating an explicit paraphrasing edit operation into an iterative simplification search algorithm. Empirically, we showed that GRS has the advantages of both approaches. GRS outperformed state-of-the-art unsupervised methods on the Newsela dataset and reduced the gap between generative and revision-based unsupervised models on the ASSET dataset. 

\bibliography{anthology,custom}
\bibliographystyle{acl_natbib}

\end{document}